# Systholic Boolean Orthonormalizer Network in Wavelet Domain for SAR Image Despeckling

Mario Mastriani

*Abstract*—We describe a novel method for removing speckle (in wavelet domain) of unknown variance from SAR images. The method is based on the following procedure: We apply 1) Bidimentional Discrete Wavelet Transform (DWT-2D) to the speckled image, 2) scaling and rounding to the coefficients of the highest subbands (to obtain integer and positive coefficients), 3) bit-slicing to the new highest subbands (to obtain bit-planes), 4) then we apply the Systholic Boolean Orthonormalizer Network (SBON) to the input bit-plane set and we obtain two orthonormal output bit-plane sets (in a Boolean sense), we project a set on the other one, by means of an AND operation, and then, 5) we apply re-assembling, and, 6) re-scaling. Finally, 7) we apply Inverse DWT-2D and reconstruct a SAR image from the modified wavelet coefficients. Despeckling results compare favorably to the most of methods in use at the moment.

## I. Introduction

A Synthetic Aperture Radar (SAR) image is affected by a particular noise called speckle in its acquisition and processing. SAR image despeckling is used to remove this multiplicative noise (of Gamma distribution [1] and [2]) while retaining as much as possible the important image features. In the recent years there has been an important amount of research on wavelet thresholding and threshold selection for SAR image despeckling, [1, 2], because wavelet provides an appropriate basis for separating noisy signal from the image signal. The motivation is that as the wavelet transform is good at energy compaction, the small coefficients are more likely due to noise and large coefficient due to important signal features [3, 4]. These small coefficients can be thresholded without affecting the significant features of the image.

Despeckling a given speckle corrupted image is a traditional problem in both biomedical and in synthetic aperture processing applications. In a SAR image, speckle manifests itself in the form of a random pixel-to-pixel variation with statistical properties similar to those of thermal noise. Due to its granular appearance in an image, speckle noise makes it very difficult to visually and automatically interpret SAR data. Therefore, speckle filtering is a critical preprocessing step for many SAR image processing tasks [1, 2], such as segmentation and classification. Many algorithms have been developed to suppress speckle noise in order to facilitate postprocessing tasks. Two types of approaches are traditionally used. The first, often referred to as multilook processing, involves the incoherent averaging of $L$ multiple looks during the generation of the SAR image. The averaging process narrows down the probability density function (pdf) of speckle and reduces the variance by a factor $L$, but this is achieved at the expense of the spatial resolution (the pixel area is increased by a factor). If the looks are not independent, such as when the Doppler bandwidth of the SAR return signal is segmented into multiple overlapping subbands, one needs to define an equivalent number of looks (ENL) [3, 4] to describe the speckle in the resultant images [1, 2].

On the other hand, the thresholding technique is the last approach based on wavelet theory to provide an enhanced approach for eliminating such noise source and ensure better interpretation of SAR images. Thresholding is a simple non-linear technique, which operates on one wavelet coefficient at a time. In its basic form, each coefficient is thresholded by comparing against threshold, if the coefficient is smaller than threshold, set to zero; otherwise it is kept or modified. Replacing the small noisy coefficients by zero and inverse wavelet transform on the result may lead to reconstruction with the essential signal characteristics and with less noise. Since the work of Donoho & Johnstone [5], there has been much research on finding thresholds, however few are specifically designed for images [1-4, 6, 7]. Unfortunately, this technique has the following disadvantages:

1) it depends on the correct election of the type of thresholding, e.g., OracleShrink, VisuShrink (soft-thresholding, hard-thresholding, and semi-soft-thresholding), Sure-Shrink, Bayesian soft thresholding, Bayesian MMSE estimation, Thresholding Neural Network (TNN), due to Zhang, NormalShrink, , etc. [3-17],
2) it depends on the correct estimation of the threshold which is arguably the most important design parameter,
3) it doesn't have a fine adjustment of the threshold after their calculation,
4) it should be applied at each level of decomposition, needing several levels, and
5) the specific distributions of the signal and noise may not be well matched at different scales.

Therefore, a new method without these constraints will represent an upgrade.

The Bidimensional Discrete Wavelet Transform and the method to reduce noise by wavelet thresholding are outlined in Section II. The SBON as a Boolean orthonormalizer and as a denoiser tool in wavelet domain is outlined in Section III. In Section IV, we discuss briefly the assessment parameters that are used to evaluate the performance of speckle reduction. In Section V, the experimental results using the proposed algorithm are presented. Finally, Section VI provides the conclusions of the paper.

## II. BIDIMENSIONAL DISCRETE WAVELET TRANSFORM

The Bidimensional Discrete Wavelet Transform (DWT-2D) [1-17] corresponds to multiresolution approximation expressions. In practice, mutiresolution analysis is carried out using 4 channel filter banks composed of a low-pass and a high-pass filter and each filter bank is then sampled at a half rate (1/2 down sampling) of the previous frequency. By repeating this procedure, it is possible to obtain wavelet transform of any order. The down sampling procedure keeps the scaling parameter constant (equal to ½) throughout successive wavelet transforms so that is benefits for simple computer implementtation. In the case of an image, the filtering is implemented in a separable way be filtering the lines and columns.

Note that [1-17] the DWT of an image consists of four frequency channels for each level of decomposition. For example, for $i$-level of decomposition we have:
LL $_{n,i}$: Noisy Coefficients of Approximation.
LH $_{n,i}$: Noisy Coefficients of Vertical Detail,
HL $_{n,i}$: Noisy Coefficients of Horizontal Detail, and
HH $_{n,i}$: Noisy Coefficients of Diagonal Detail.

The LL part at each scale is decomposed recursively, as illustrated in Fig. 1.

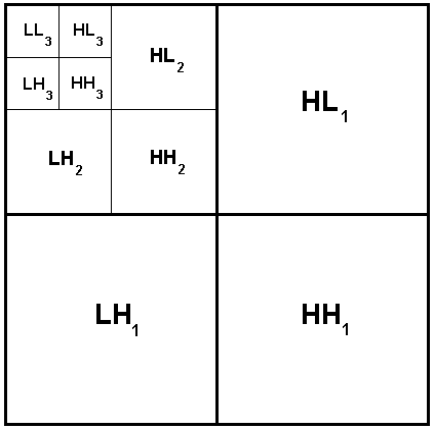

Fig. 1 Data preparation of the image. Recursive decomposition of LL parts.

To achieve space-scale adaptive noise reduction, we need to prepare the 1-D coefficient data stream which contains the space-scale information of 2-D images. This is somewhat similar to the "zigzag" arrangement of the DCT (Discrete Cosine Transform) coefficients in image coding applications [18]. In this data preparation step, the DWT-2D coefficients are rearranged as a 1-D coefficient series in spatial order so that the adjacent samples represent the same local areas in the original image. An example of the rearrangement of an 8-by-8 transformed image is shown in Fig. 2, which will be referred to as a 1D space-scale data stream.

Each number in Fig. 2 represents the spatial order of the 2D coefficient at that position corresponding to Fig. 1.

| 64 | 63 | 15 | 45 | 3 | 9 | 33 | 39 |
|----|----|----|----|----|----|----|----|
| 62 | 61 | 30 | 60 | 6 | 12 | 36 | 42 |
| 14 | 44 | 13 | 43 | 18 | 24 | 48 | 54 |
| 29 | 59 | 28 | 58 | 21 | 27 | 51 | 57 |
| 2 | 8 | 32 | 38 | 1 | 7 | 31 | 37 |
| 5 | 11 | 35 | 41 | 4 | 10 | 34 | 40 |
| 17 | 23 | 47 | 53 | 16 | 22 | 46 | 52 |
| 20 | 26 | 50 | 56 | 19 | 25 | 49 | 55 |

Fig. 2 Data preparation of the image. Spatial order of 2-D coefficients.

### A. Wavelet Noise Thresholding

The wavelet coefficients calculated by a wavelet transform represent change in the image at a particular resolution. By looking at the image in various resolutions it should be possible to filter out noise. At least in theory. However, the definetion of noise is a difficult one. In fact, "one person's noise is another's signal". In part this depends on the resolution one is looking at. One algorithm to remove Gaussian white noise is summarized by D. L. Donoho and I. M. Johnstone [5], and synthesized in Fig. 3.

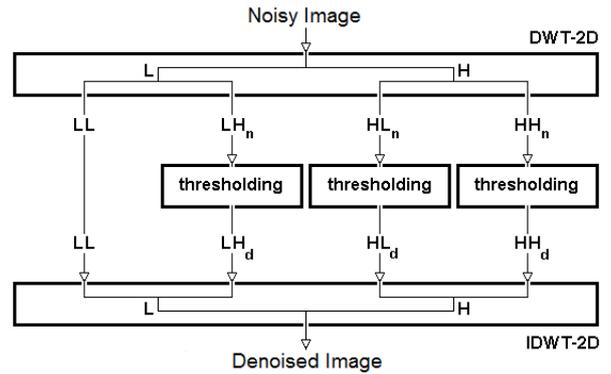

Fig. 3 Thresholding Techniques

The algorithm is:
1) Calculate a wavelet transform and order the coefficients by increasing frequency. This will result in an array containing the image average plus a set of coefficients of length 1, 2, 4, 8, etc. The noise threshold will be calculated on the highest frequency coefficient spectrum (this is the largest spectrum).

2) Calculate the *median absolute deviation* on the largest coefficient spectrum. The median is calculated from the absolute value of the coefficients. The equation for the median absolute deviation is shown below:

$$\delta_{mad} = \frac{median(|C_{n,i}|)}{0.6745} \quad (1)$$

where $C_{n,i}$ may be $LH_{n,i}$, $HL_{n,i}$, or $HH_{n,i}$ for *i*-level of decomposition. The factor 0.6745 in the denominator rescales the numerator so that $\delta_{mad}$ is also a suitable estimator for the standard deviation for Gaussian white noise [1, 2, 5, 7-12].

3) For calculating the noise threshold $\lambda$ we have used a modified version of the equation that has been discussed in papers by D. L. Donoho and I. M. Johnstone. The equation is:

$$\lambda = \delta_{mad}\sqrt{2log[N]} \quad (2)$$

where N is the number of pixels in the subimage, i.e., HL, LH or HH.

4) Apply a thresholding algorithm to the coefficients. There are two popular versions:
4.1. Hard thresholding. Hard thresholding sets any coefficient less than or equal to the threshold to zero, see Fig. 4(a).

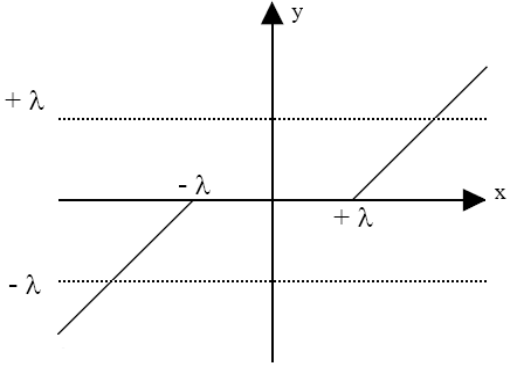

Fig. 4(a) Soft-Thresholfing

where *x* may be $LH_{n,i}$, $HL_{n,i}$, or $HH_{n,i}$, *y* may be $HH_{d,i}$: Denoised Coefficients of Diagonal Detail,
$HL_{d,i}$: Denoised Coefficients of Horizontal Detail,
$LH_{d,i}$: Denoised Coefficients of Vertical Detail,
for *i*-level of decomposition.

The respective code is:

```
for row = 1:N^(1/2)
  for column = 1:N^(1/2)
    if |C_n,i[row][column]| <= λ,
      C_n,i[row][column] = 0.0;
    end
  end
end
```

4.2. Soft thresholding. Soft thresholding sets any coefficient less than or equal to the threshold to zero, see Fig. 4(b). The threshold is subtracted from any coefficient that is greater than the threshold. This moves the image coefficients toward zero.

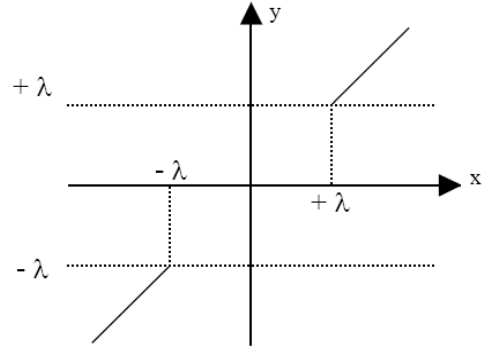

Fig. 4(b): Hard-Thresholfing

The respective code is:

```
for row = 1:N^(1/2)
  for column = 1:N^(1/2)
    if |C_n,i[row][column]| <= λ,
      C_n,i[row][column] = 0.0;
    else
      C_n,i[row][column] = C_n,i[row][column] - λ;
    end
  end
end
```

III. SYNTHOLIC BOOLEAN ORTHONORMALIZER NETWORK

The SBON was introduced by Mastriani [19] as a Boolean Orthonormalization Process (*BOP*) to convert a non-orthonormal Boolean basis, i.e., a set of non-orthonormal binary vectors (in a Boolean sense) to an orthonormal Boolean basis, i.e., a set of orthonormal binary vectors (in a Boolean sense). The BOP algorithm has a lot of fields of applications, e.g.: Steganography, Hopfield Networks, Boolean Correlation Matrix Memories [20], Bi-level image processing, lossy compression, iris, fingerprint and face recognition, improving edge detection and image segmentation, among others. That is to say, all those applications that need orthonormality in a Boolean sense. It is important to mention that the BOP is an extremely stable and fast algorithm.

*A. Orthonormality in a Boolean Sense*

Given a set of binary vectors $\mathbf{u}_k = [\ u_{k1}, u_{k2}, \ldots, u_{kp}\ ]^T$ (where $k = 1, 2, \ldots, q$, and $[.]^T$ means transpose of $[.]$), *they are orthonormals in a Boolean sense*, if they satisfy the following pair of conditions:

$$u_k \wedge u_j = u_k = u_j \text{ if } k = j \quad (3.1)$$

and

$$u_k \wedge u_j = 0 \text{ if } k \neq j \quad (3.2)$$

where $0 = [\,0\,,0\,,..., 0\,]^T$, and the term $u_k \wedge u_j$ represents the *AND operation between each element of the binary vectors* $u_k$ and $u_j$, i.e.,

$$u_k \wedge u_j = [u_{k1} \wedge u_{j1}, u_{k2} \wedge u_{j2}, \bullet\bullet\bullet, u_{kp} \wedge u_{jp}]^T \quad (4)$$

*B. Boolean Orthonormalization Process (BOP)*

Given a set of key binary vectors that are nonorthonormal (in a Boolean sense), we may use a *preprocessor* to transform them into an orthonormal set (in a Boolean sense); the preprocessor is designed to perform a *Boolean orthonormalization* on the key binary vectors prior to association. This form of transformation is described below, maintaining a one-to-one correspondence between the input (key) binary vectors $v_1, v_2, ..., v_N$, the resulting orthonormal binary vectors $u_1, u_2, ..., u_N$, and the residual binary vectors $s_1, s_2, ..., s_N$.

1) Version 1: As it is shown in the Fig. 5

$$s_i = v_i \veebar u_i \quad \forall\ i,\ \text{with}\ s_1 = 0 \quad (5)$$

$$u_i \wedge s_i = 0 \quad \forall\ i \quad (6)$$

where $\veebar$ represents the XOR operation. Eq.(6) represents the orthogonality principle in a Boolean sense [19].

$$u_i \wedge u_k = 0 \quad \forall\ i \neq k \quad (7)$$

$$s_{i,j} \leq v_{i,j} \quad \forall\ i, j \quad (8)$$

*Algorithm:*

$u_j = v_j \quad \forall\ j$

$u_j = u_j \veebar (v_j \wedge u_i),\ i \in [1, j\text{-}1],\ j \in [1, N]$

2) Version 2: As it is shown in the Fig. 6

$$u_i = v_i \veebar s_i \quad \forall\ i,\ \text{with}\ u_1 = v_1,\ \text{because}\ s_1 = 0 \quad (9)$$

*Algorithm:*

$s_j = 0 \quad \forall\ j$

$s_j = s_j \vee (v_j \wedge (v_i \veebar s_i)),\ i \in [1, j\text{-}1],\ j \in [1, N]$

*C. SBON in wavelet domain for SAR image despeckling*

The new method of SAR image despeckling can be represented by the Fig. 7, according to the following algorithm:

We apply
1) DWT-2D to the speckled SAR image,

2) scaling and rounding to the coefficients of the highest subbands (for to obtain integer and positive coefficients),

3) bit-slicing to the new highest subbands (for to obtain bit-planes), see Fig. 8

4) SBON to the input bit-plane set and we obtain two orthonormal output bit-plane sets (in a Boolean sense), see Fig. 9, we project a set on the other one, by means of an AND operation, and then, see Fig. 7

5) re-assembling, and, see Fig. 8

6) re-scaling, and

7) Inverse DWT-2D and reconstruct a SAR image from the modified wavelet coefficients.

## IV. ASSESSMENT PARAMETERS

In this work, the assessment parameters that are used to evaluate the performance of speckle reduction are
1) for simulated speckled images: Signal-to-Noise Ratio [18], and Pratt's figure of Merit [21, 22].

2) for real speckled images: Noise Variance, Mean Square Difference, Noise Mean Value, Noise Standard Deviation, Equivalent Number of Looks, Deflection Ratio [1-4], and Pratt's figure of Merit [21, 22].

*A. Noise Mean Value (NMV), Noise Variance (NV), Mean Square Error (MSE), and Signal-to-Noise Ratio (SNR)*

The SNR is defined as the ratio of the variance of the noise-free signal $I$ to the MSE between the noise-free signal and the denoised signal $\hat{I}$ [18]. The formulas for the *NMV and NV* calculation are

$$NMV = \frac{\sum_{r,c} \hat{I}(r,c)}{R*C} \quad (10)$$

$$NV = \frac{\sum_{r,c}\left(\hat{I}(r,c) - NMV\right)^2}{R*C} \quad (11)$$

where *R-by-C* pixels is the size of the despeckled image $\hat{I}$. On the other hand, the estimated noise variance is used to determine the amount of smoothing needed for each case for all filters. *NV* determines the contents of the speckle in the image. A lower variance gives a "cleaner" image as more speckle is reduced, although, it not necessarily depends on the intensity. The formulas for the *MSE and SNR are*

$$MSE = \frac{\sum_{r,c}(I(r,c) - \hat{I}(r,c))^2}{R*C} \quad (12)$$

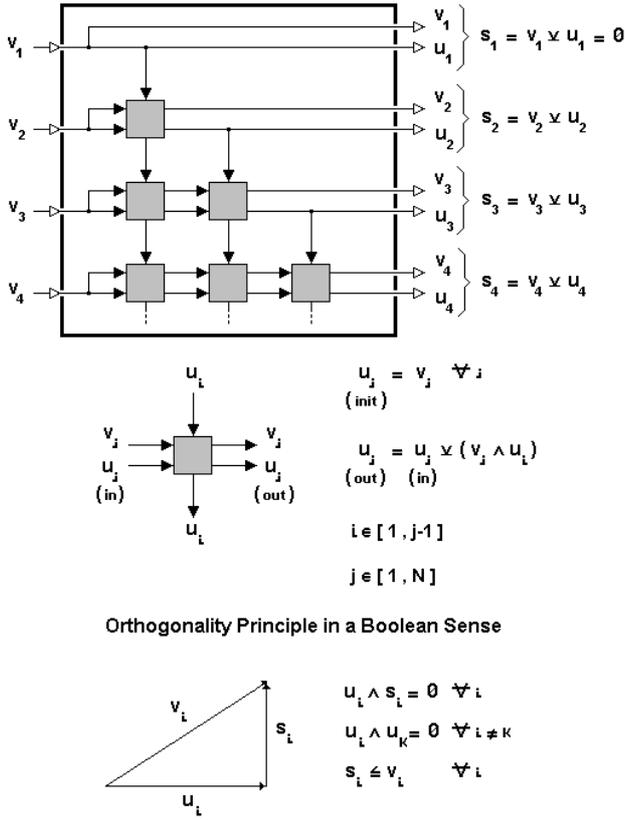

Fig. 5 SBON, version 1

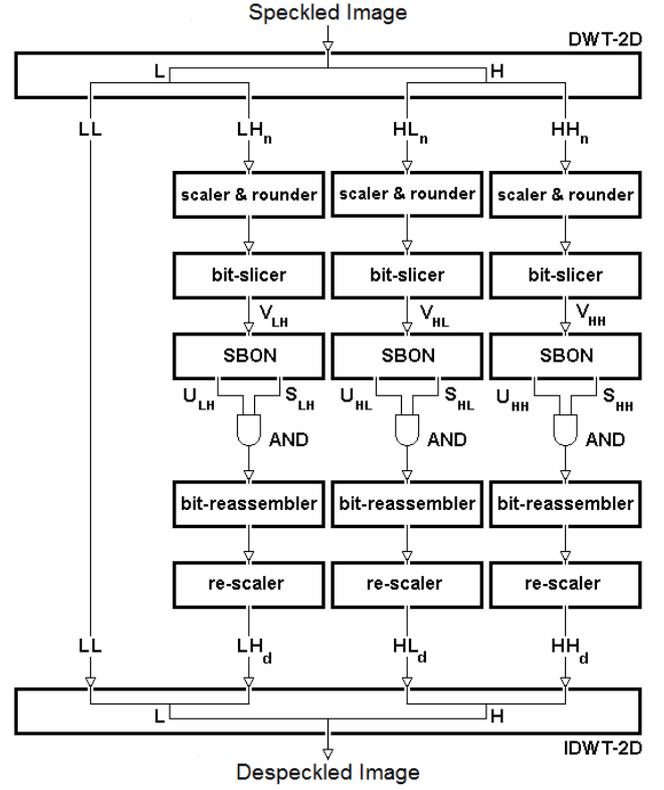

Fig. 7 SAR despeckling

$$SNR = 10\ \log_{10}\left(\frac{NV}{MSE}\right) \qquad (13)$$

### B. Noise Standard Deviation (NSD)

The formula for the *NSD* calculation is

$$NSD = \sqrt{NV} \qquad (14)$$

### C. Mean Square Difference (MSD)

*MSD* indicates average square difference of the pixels throughout the image between the original image (with speckle) $I_s$ and $\hat{I}$, as shown in Fig. 7. A lower *MSD* indicates a smaller difference between the original (with speckle) and despeckled image. This means that there is a significant filter performance. Nevertheless, it is necessary to be very careful with the edges. The formula for the *MSD* calculation is

$$MSD = \frac{\sum_{r,c}(I_s(r,c) - \hat{I}(r,c))^2}{R*C} \qquad (15)$$

### D. Equivalent Numbers of Looks (ENL)

Another good approach of estimating the speckle noise level in a SAR image is to measure the *ENL* over a uniform image region [3, 4]. A larger value of *ENL* usually corresponds to a better quantitative performance.

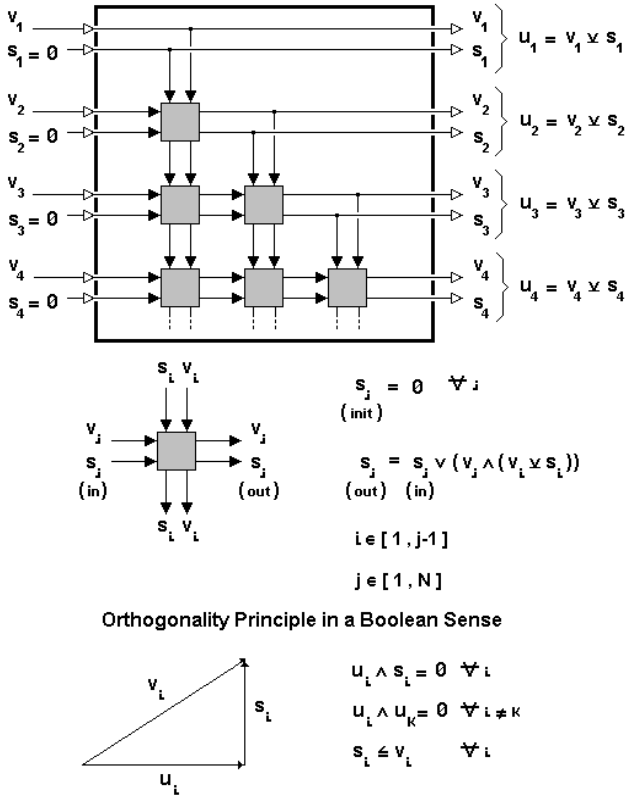

Fig. 6 SBON, version 2

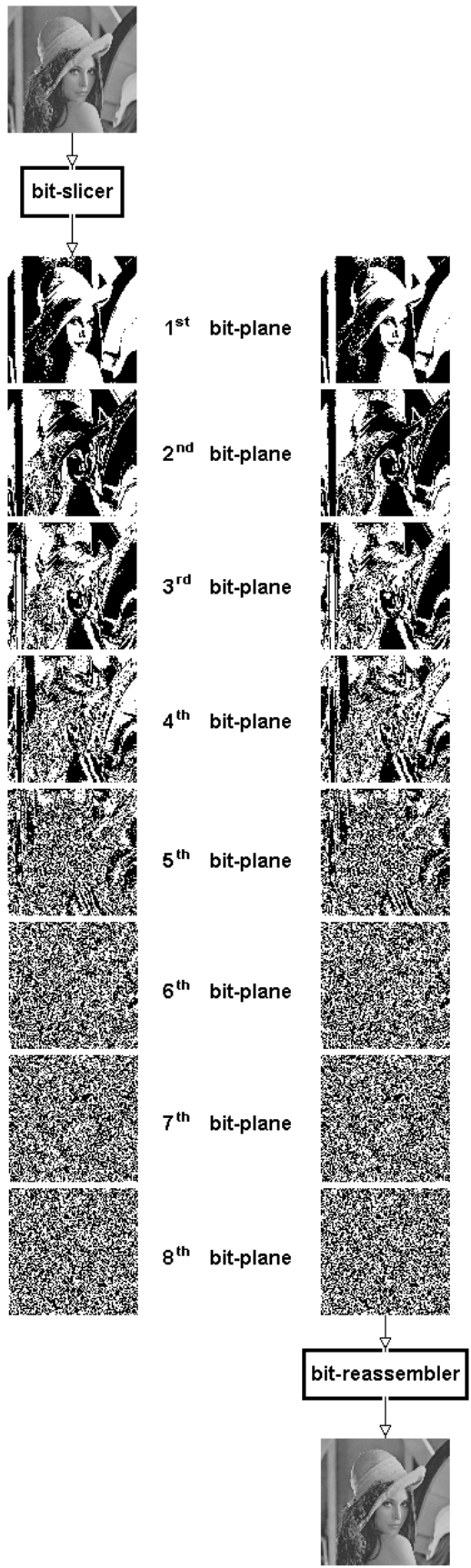

Fig. 8 Lena's bit-slicing and re-assembling.

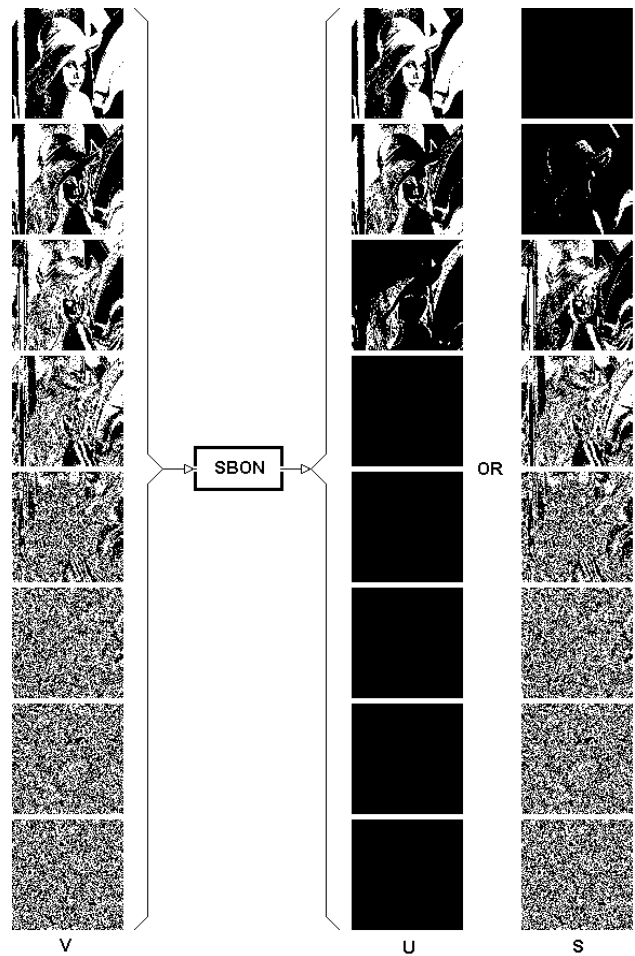

Fig. 9 SBON application on Lena's bit-plane set

The value of *ENL* also depends on the size of the tested region, theoretically a larger region will produces a higher *ENL* value than over a smaller region but it also tradeoff the accuracy of the readings.

Due to the difficulty in identifying uniform areas in the image, we proposed to divide the image into smaller areas of 25x25 pixels, obtain the *ENL* for each of these smaller areas and finally take the average of these *ENL* values. The formula for the *ENL* calculation is

$$ENL = \frac{NMV^2}{NSD^2} \qquad (16)$$

The significance of obtaining both *MSD* and *ENL* measurements in this work is to analyze the performance of the filter on the overall region as well as in smaller uniform regions.

### E. Deflection Ratio (DR)

A fourth performance estimator that we used in this work is the *DR* proposed by H. Guo et al (1994), [4]. The formula for the deflection calculation is

$$DR = \frac{1}{R*C}\sum_{r,c}\left(\frac{\hat{I}(r,c)-NMV}{NSD}\right) \quad (17)$$

The ratio *DR* should be higher at pixels with stronger reflector points and lower elsewhere. In H. Guo *et al*'s paper, this ratio is used to measure the performance between different wavelet shrinkage techniques. In this paper, we apply the ratio approach to all techniques after despeckling in the same way [1, 2].

### F. Pratt's figure of merit (FOM)

To compare edge preservation performances of different speckle reduction schemes, we adopt the Pratt's figure of merit [21, 22] defined by

$$FOM = \frac{1}{max\{\hat{N},N_{ideal}\}}\sum_{i=1}^{\hat{N}}\frac{1}{1+d_i^2\alpha} \quad (18)$$

Where $\hat{N}$ and $N_{ideal}$ are the number of detected and ideal edge pixels, respectively, $d_i$ is the Euclidean distance between the *i*th detected edge pixel and the nearest ideal edge pixel, and α is a constant typically set to 1/9. *FOM* ranges between *0* and *1*, with unity for ideal edge detection.

## V. EXPERIMENTAL RESULTS

### A. For Images with Simulated Speckle

Here, we present a set of experimental results using the SBON technique in standard 256-by-256 Lena image. The other methods against which we assess the performance of the proposed speckle filter include the following: the Bayesian soft thresholding technique proposed in [6]; the Bayesian MMSE estimation technique using the Gaussian mixture density model referenced in [1]; the refined Lee filter [1, 2]; and the Wiener filter [18]. Besides, Fig. 10 shows the noisy image used in this experiment, and the filtered images. Table I shows the assessment parameters vs. 4 filters for Fig. 10.

Table I. Assessment Parameters vs. Filters for Fig. 10.

| Filters | Assessment Parameters | |
|---|---|---|
| | SNR | FOM |
| Noisy observation | 0.5432 | 0.37486 |
| Bayes soft thresholding | 0.8976 | 0.42311 |
| Bayes MMSE estimation | 0.8645 | 0.42387 |
| Refined Lee | 0.8712 | 0.42867 |
| Wiener | 0.8809 | 0.42111 |
| SBON | 0.9989 | 0.46132 |

### B. For Images with Real Speckle

Here, we present a set of experimental results using one ERS SAR Precision Image (PRI) standard of Buenos Aires area. For statistical filters employed along, i.e., Median, Lee, Kuan, Gamma-Map, Enhanced Lee, Frost, Enhanced Frost [1, 4], Wiener [18], DS [8] and Enhanced DS (EDS) [22], we use a homomorphic speckle reduction scheme [22], with 3-by-3, 5-by-5 and 7-by-7 kernel windows. Besides, for Lee, Enhanced Lee, Kuan, Gamma-Map, Frost and Enhanced Frost filters the damping factor is set to 1 [1-4].

Fig. 11 shows a noisy image used in the experiment from remote sensing satellite ERS-2, with a 242-by-242 (pixels) by 65536 (gray levels); and the filtered images, processed by using VisuShrink (Hard-Thresholding), BayesShrink, Normal-Shrink, SUREShrink, and NeuralShrink techniques respectively, see Table II.

All the wavelet-based techniques used Daubechies 1 wavelet basis and 1 level of decomposition (improvements were not noticed with other basis of wavelets) [6, 7, 21]. Besides, Fig. 11 summarizes the edge preservation performance of the SBON technique vs. the rest of the shrinkage techniques with a considerably acceptable computational complexity.

Table II shows the assessment parameters vs. 19 filters for Fig. 11, where En-Lee means Enhanced Lee Filter, En-Frost means Enhanced Frost Filter, Non-log SWT means Non-logarithmic Stationary Wavelet Transform Shrinkage [4], Non-log DWT means Non-logarithmic DWT Shrinkage [3-7, 13, 15], VisuShrink (HT) means Hard-Thresholding, (ST) means Soft-Thresholding, and (SST) means Semi-ST [3, 5, 6, 13, 15, 16].

We compute and compare the NMV and NSD over six different homogeneous regions in our SAR image, before and after filtering, for all filters.

The SBON has obtained the best mean preservation and variance reduction, as shown in Table II. Since a successful speckle reducing filter will not significantly affect the mean intensity within a homogeneous region, SBON demonstrated to be the best in this sense too. The quantitative results of Table II show that the SBON technique can eliminate speckle without distorting useful image information and without destroying the important image edges. In fact, the SBON outperformed the conventional and no conventional speckle reducing filters in terms of edge preservation measured by Pratt's figure of merit [21, 22], as shown in Table II.

On the other hand, all filters was applied to complete image, for Figure 10 (256-by-256) pixels and Figure 11 (256-by-256) pixels, and all the filters were implemented in MATLAB® (Mathworks, Natick, MA) R2012a on a Alienware M17X10 Intel® Core™ i7 CPU Q740 @ 1.73 GHz x 2, with 6 GB RAM, on a 64-bit Operating Systems, Windows 7 Ultimate.

## VI. Conclusion

In this paper, we developed a new application of SBON, this time to SAR speckled images, which combines a non-linear filtering with shrinkage methods. We reused the same architectture that was employed in microarrays to serve as the shrinkage function of DWT-2D for speckled SAR images. Unlike the standard techniques, SBON has a lower computational cost. By using this new shrinkage function, we are introducing (first time) a bit level method for wavelet denoising/despeckling processes.

We then discussed the high performance solution of the SBON in the SNR and FOM sense. It is proved that there is at most two top application for the new shrinkage method in the wavelet domain (microarrays denoising and SAR despeckling). The general optimal performance of SBON is analyzed and compared to the linear speckle reduction method. It is shown that the shrinkage speckle reduction methods are more effecttive than linear methods when the signal energy concentrates on few coefficients in the transform domain. Besides, considerably increased deflection ratio strongly indicates improvement in detection performance. Finally, the method is computationally efficient and can significantly reduce the speckle while preserving the resolution of the original image, and avoiding several levels of decomposition and block effect.

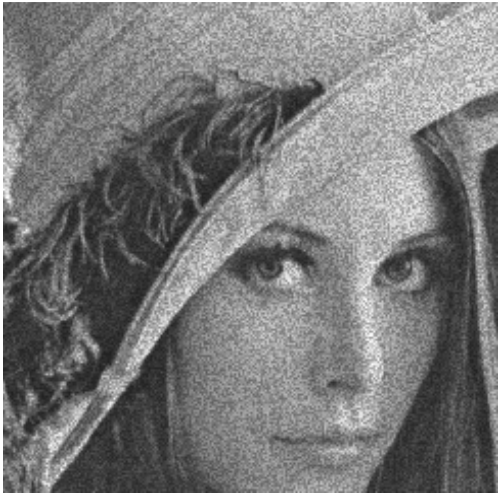
original

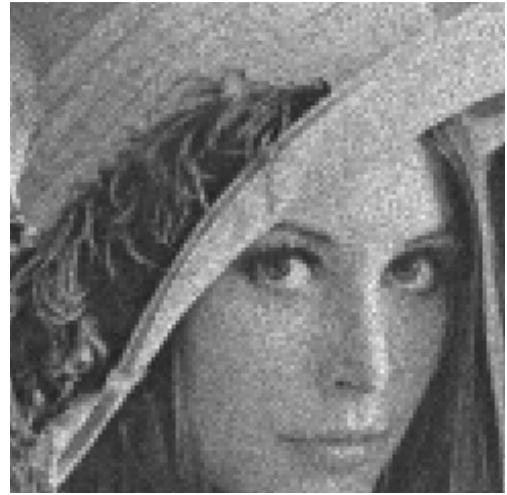
OracleShrink

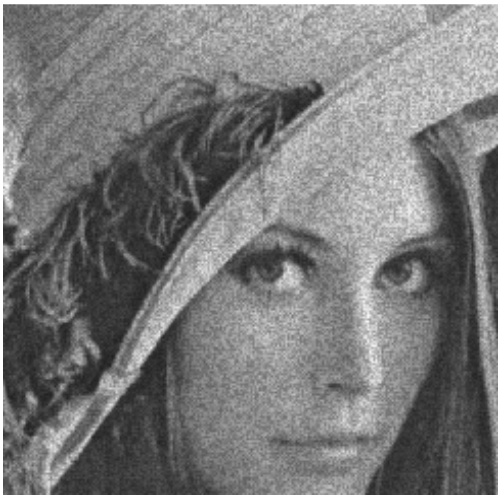
NormalShrink

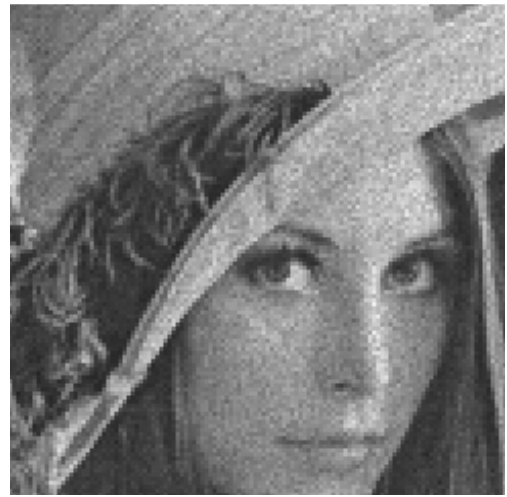
SUREShrink

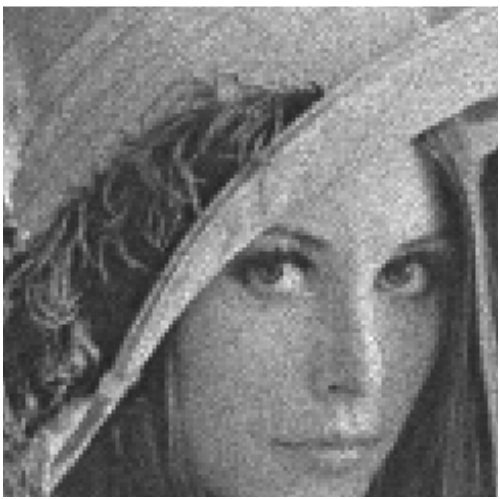
BayesShrink

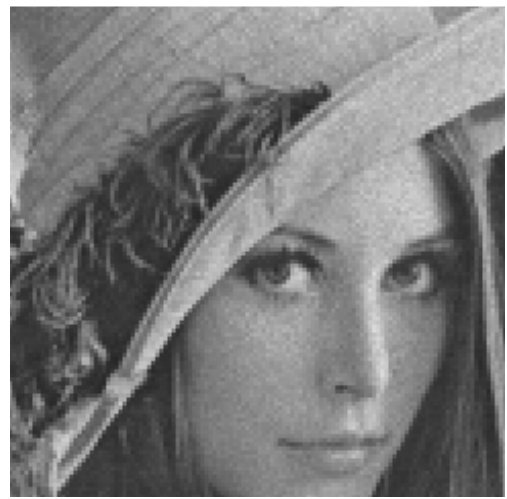
SBON

Fig. 10 Original speckled and despeckled images.

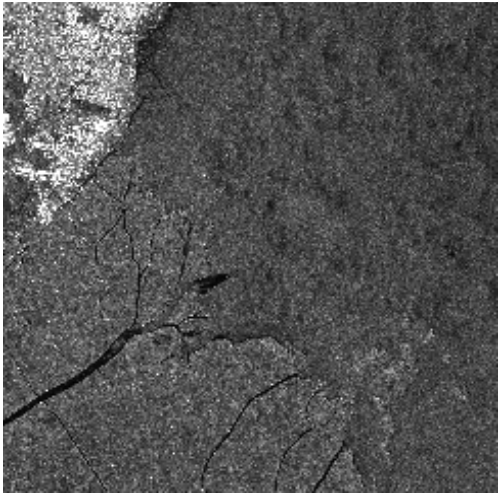
original

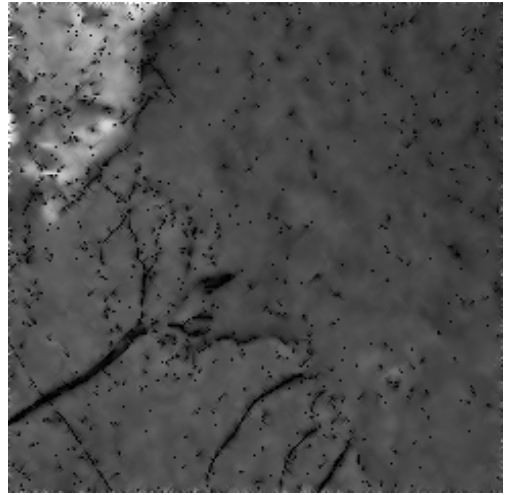
OracleShrink

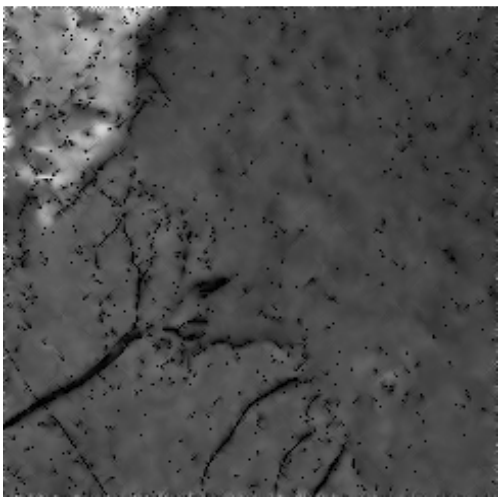
NormalShrink

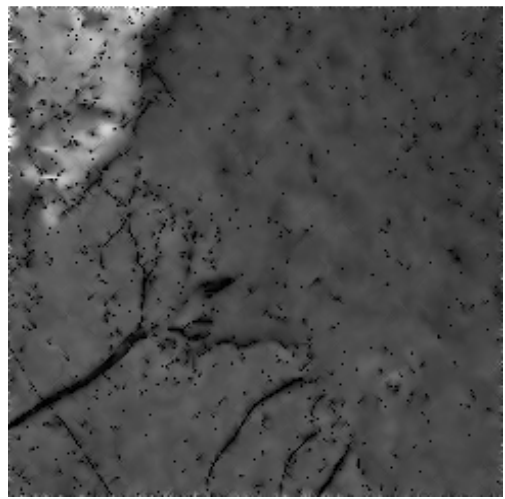
SUREShrink

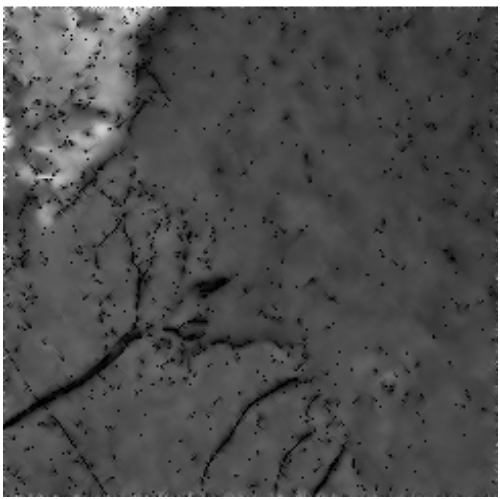
BayesShrink

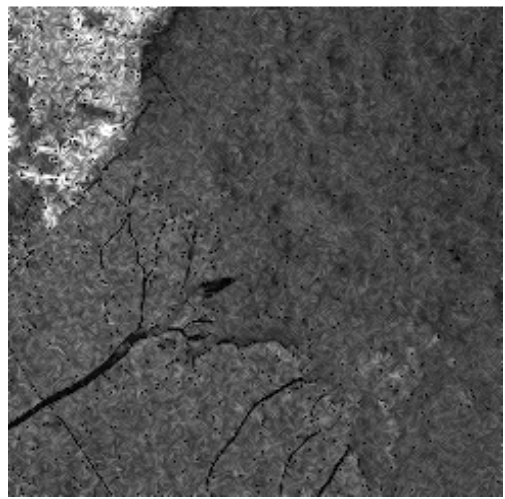
SBON

Fig. 11 Original and despeckled images.

TABLE II
ASSESSMENT PARAMETERS VS. FILTERS FOR FIGURE 11

| Filter | Assessment Parameter | | | | | |
|---|---|---|---|---|---|---|
| | MSD | NMV | NSD | ENL | DR | FOM |
| Original Speckled Image | - | 90.0890 | 43.9961 | 11.0934 | 2.5580e-017 | 0.3027 |
| En-Frost | 564.8346 | 87.3245 | 40.0094 | 16.3454 | 4.8543e-017 | 0.4213 |
| En-Lee | 532.0006 | 87.7465 | 40.4231 | 16.8675 | 4.4236e-017 | 0.4112 |
| Frost | 543.9347 | 87.6463 | 40.8645 | 16.5331 | 3.8645e-017 | 0.4213 |
| Lee | 585.8373 | 87.8474 | 40.7465 | 16.8465 | 3.8354e-017 | 0.4228 |
| Gamma | 532.9236 | 87.8444 | 40.6453 | 16.7346 | 3.9243e-017 | 0.4312 |
| Kuan | 542.7342 | 87.8221 | 40.8363 | 16.9623 | 3.2675e-017 | 0.4217 |
| Median | 614.7464 | 85.0890 | 42.5373 | 16.7464 | 2.5676e-017 | 0.4004 |
| Wiener | 564.8346 | 89.8475 | 40.3744 | 16.5252 | 3.2345e-017 | 0.4423 |
| DS | 564.8346 | 89.5353 | 40.0094 | 17.8378 | 8.5942e-017 | 0.4572 |
| EDS | 564.8346 | 89.3232 | 40.0094 | 17.4242 | 8.9868e-017 | 0.4573 |
| VisuShrink (ST) | 855.3030 | 88.4311 | 32.8688 | 39.0884 | 7.8610e-016 | 0.4519 |
| VisuShrink (HT) | 798.4422 | 88.7546 | 32.9812 | 38.9843 | 7.7354e-016 | 0.4522 |
| VisuShrink (SST) | 743.9543 | 88.4643 | 32.9991 | 37.9090 | 7.2653e-016 | 0.4521 |
| SureShrink | 716.6344 | 87.9920 | 32.8978 | 38.3025 | 2.4005e-015 | 0.4520 |
| OracleShrink | 732.2345 | 88.5233 | 33.3124 | 36.8464 | 6.7354e-016 | 0.4576 |
| BayesShrink | 724.0867 | 88.9992 | 36.8230 | 36.0987 | 1.0534e-015 | 0.4581 |
| NormalShrink | 300.2841 | 86.3232 | 43.8271 | 11.2285 | 1.5783e-016 | 0.4577 |
| TNN | 341.3989 | 87.1112 | 39.4162 | 16.4850 | 1.0319e-015 | 0.4588 |
| SBON | 334.7461 | 86.0014 | 38.8903 | 39.3564 | 2.9475e-015 | 0.4629 |